\documentclass{article}
\usepackage{spconf,amsmath,graphicx,amsfonts,amssymb}
\usepackage{enumitem}
\newlist{steps}{enumerate}{1}
\usepackage{diagbox}

\usepackage{multirow}
\usepackage{amssymb}
\usepackage{caption}
\usepackage{subcaption}
\usepackage{hyperref}
\usepackage[normalem]{ulem}
\usepackage{url}
\usepackage{cite}

\title{Revisiting IPA-based Cross-lingual Text-to-speech}

\name{Haitong Zhang, Haoyue Zhan, Yang Zhang, Xinyuan Yu, Yue Lin}
\address{Netease Games AI Lab, China \\
\{zhanghaitong01, zhanhaoyue, zhangyang09, yuxinyuan, gzlinyue\}@corp.netease.com}
\begin{document}
\ninept
\maketitle
\begin{abstract}
International Phonetic Alphabet (IPA) has been widely used in cross-lingual text-to-speech (TTS) to achieve cross-lingual voice cloning (CL VC). However, IPA itself has been understudied in cross-lingual TTS. In this paper, we report some empirical findings of building a cross-lingual TTS model using IPA as inputs. Experiments show that the way to process the IPA  and suprasegmental sequence has a negligible impact on the CL VC performance. Furthermore, we find that using a dataset including one speaker per language to build an IPA-based TTS system would fail CL VC since the language-unique IPA and tone/stress symbols could leak the speaker information. In addition, we experiment with different combinations of speakers in the training dataset to further investigate the effect of the number of speakers on the CL VC performance.
\end{abstract}


\section{Introduction}
Recently, text-to-speech (TTS) has witnessed a rapid development in synthesizing mono-language speech using sequence-to-sequence models \cite{shen2018natural, DBLP:conf/iclr/PingPGAKNRM18, ren2020fastspeech} and high-fidelity neural vocoders \cite{oord2018parallel, 48585,kong2020hifi}. Meanwhile, researchers have begun to study cross-lingual TTS, whose main challenge may lie in disentangling language attributes from speaker identities to achieve cross-lingual voice cloning (CL VL). 

Normally, multi-lingual speech from the multi-lingual speaker is required to build a TTS system that can perform CL VL \cite{Traber99frommultilingual}. However, it is hard to find a speaker who is proficient in multiple languages and has smooth articulation across different languages \cite{DBLP:conf/interspeech/XueSXXW19}. Thus, researchers have started to study building cross-lingual TTS systems using mono-lingual data.

Researchers initially investigated code-switched TTS by sharing the HMM states across different languages \cite{1415035,4730269, 5153557}  , formant mapping based frequency warping \cite{He2012TurningAM}, and using a unified phone set for multiple languages \cite{Chandu2017Speech}.

More recently, researchers have started to investigate sequence-to-sequence based cross-lingual TTS. \cite{8682927} proposes to use separate encoders to handle alphabet inputs of different languages. \cite{DBLP:conf/interspeech/XueSXXW19}
 adopts the pretrain-and-finetune method to build a cross-lingual TTS system using mono-lingual data.  \cite{48331, liu2020tone, xin2020cross} use a gradient reversal layer to disentangle speaker information from the textual encoder. \cite{Nekvinda2020OneMM} uses meta-learning to improve the cross-lingual performance. \cite{Nekvinda2020OneMM} uses graphemes as the input representations, while \cite{8682674} proposes to use bytes as model inputs, resulting in synthesizing fluent code-switched speech; but the voice switches for different languages. 
\cite{zhan2021improve} compares the CL VL performance between language-dependent phoneme and language-independent phoneme (IPA) based multi-lingual TTS systems.
 \cite{cao2020code} uses bilingual phonetic posteriorgram (PPG) to achieve code-switched TTS.

\begin{figure}
    \centering
    \includegraphics[width=0.45\textwidth]{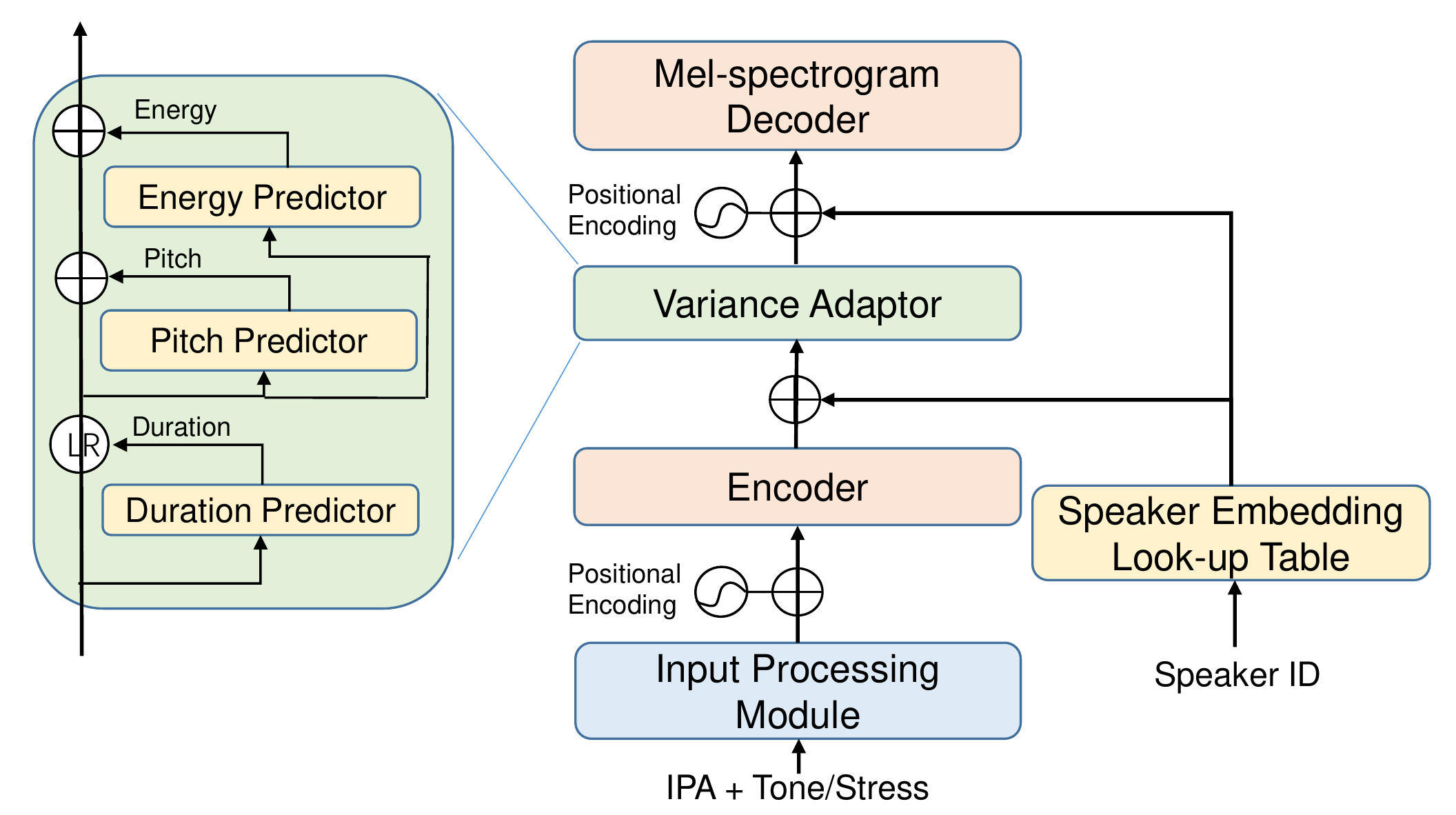}
    \caption{\emph{Model structure studied.}}
    \label{fig:model}
\end{figure}

\subsection{The contribution}
Although IPA has been widely used in cross-lingual TTS \cite{48331, zhan2021improve}, IPA itself has been understudied in cross-lingual TTS. In this paper, we conduct an empirical study of IPA in cross-lingual TTS, with an attempt to answer the following questions:

     
    

\begin{itemize}
    \item Does the way to process IPA and suprasegmental sequences have a significant impact on the CL VL performance?
    \item Is monolingual data from only two speakers (one speaker per language) sufficient to achieve a promising CL VL performance in the IPA-based cross-lingual model? 
    \item What is the impact of the number of speakers per language on the CL VL performance?
\end{itemize}

To answer these questions, we conduct a performance comparison between two IPA processing modules in the non-autoregressive TTS model. Besides, we analyze the cross-lingual TTS model trained with only one speaker per language by devising two input perturbation methods and compare the number of speakers per language to analyze its effect on the CL VL performance.

\section{Framework}
\subsection{Model architecture}
The core of the framework is Fastspeech2 model \cite{ren2020fastspeech}, a transformer-based non-autoregressive TTS model. The model mainly consists of an encoder, a variance adaptor, and a mel-spectrogram decoder. The encoder converts the input embedding sequence into the hidden sequence, and then the variance adaptor
adds different variance information such as duration, pitch, and energy into the hidden sequence; finally, the mel-spectrogram decoder converts the adapted hidden sequence into mel-spectrogram sequence in parallel. To support multi-speaker TTS, we extract the speaker embedding from the speaker embedding look-up table and place it at two positions: 1) adding to the encoder output and 2) adding to the decoder input. The overall structure is illustrated in Fig. \ref{fig:model}.

\subsection{Input processing module}
\begin{figure}[t]
\centering
\includegraphics[width=0.45\textwidth]{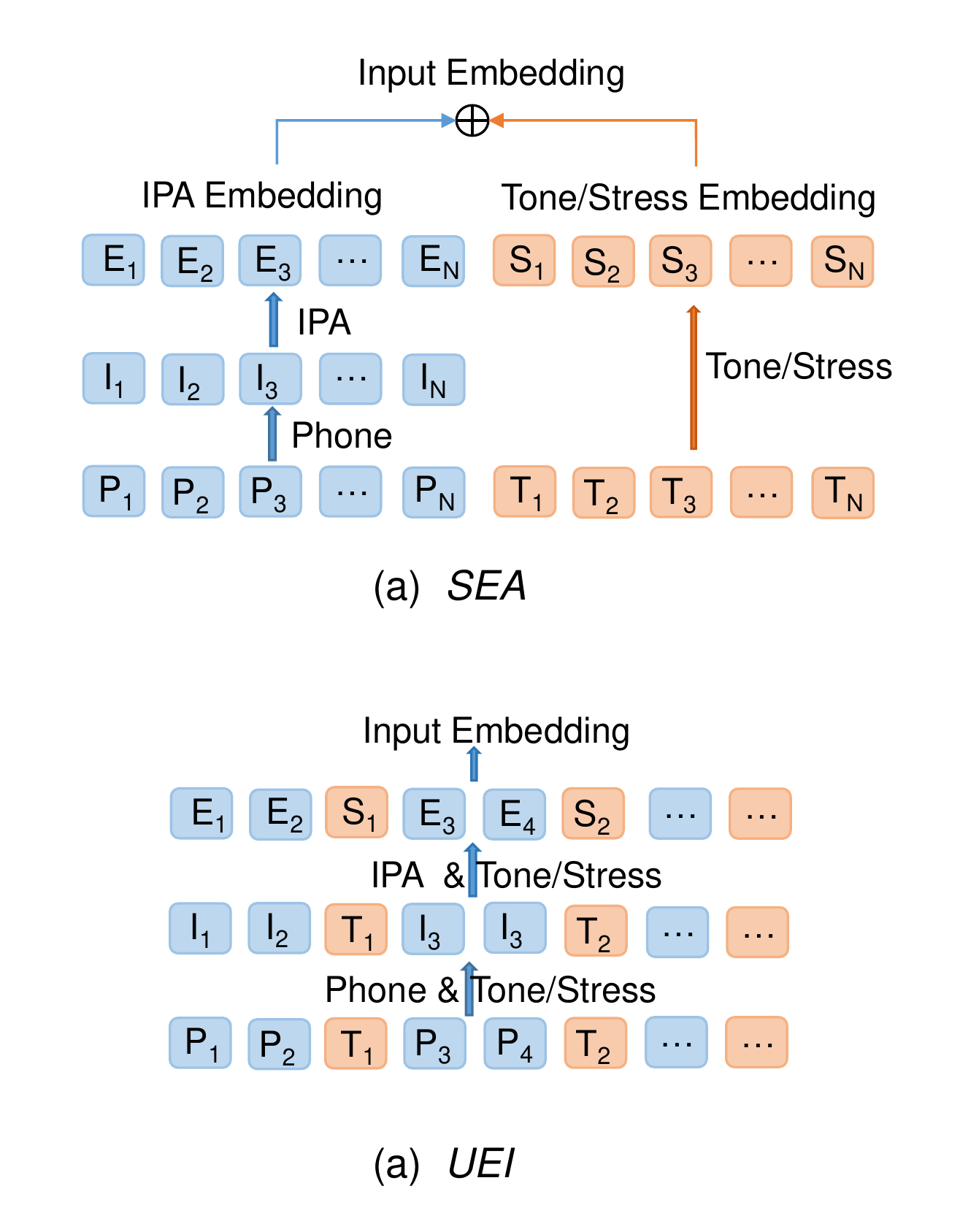}
\caption{\emph{Examples of the input processing modules, where $\oplus$ denotes element-wise addition. Prosody symbols are ignored here for brevity.}}
\label{fig:IPM}
\end{figure}

The input of the cross-lingual model usually includes IPA and suprasegmental, including tone/stress. To investigate whether the way to process them has an impact on the CL VL performance, we consider two different processing modules: 1) SEA:  use \textbf{S}eparate \textbf{E}mbedding sequences for IPA and tone/stress, then \textbf{A}dd two embedding sequences to form the final input embedding; 2) UEI: use \textbf{U}nified \textbf{E}mbedding for IPA and tone/stress, then take each embedding as an \textbf{I}ndependent input in the final input embedding. We illustrate these two processing modules in Fig. \ref{fig:IPM}.

\section{Experimental setup}
\subsection{Data} \label{data}

 
 In this paper, we implement experiments on Chinese (Mandarin) and English. We include two datasets in this paper. Dataset1 consists of the monolingual speech from two female speakers: a Chinese speaker \cite{CSMSC} and an English speaker \cite{ljspeech17}. Each speaker has roughly ten hours of speech. We use 200 utterances for evaluation and the rest for training. Besides Dataset1, Dataset2 includes monolingual data from four additional speakers (one female and male from \cite{bakhturina2021hi} and one female and male from our proprietary speech corpus). Each speaker has about 5 to 10 hours of monolingual data.

\subsection{Implementation details}


We use G2P converter to convert text sequence into language-dependent phoneme sequence, and then convert them into IPA \footnote{ \url{https://en.wikipedia.org/wiki/International_Phonetic_Alphabet} } and tone/stress sequence. We include five tones for Chinese Mandarin and two stresses (primary and secondary stress) for English. A special symbol is used when there is no tone and stress. We also include a word boundary symbol in the input sequence. We use Montreal forced alignment (MFA) \cite{mcauliffe2017montreal} tool to extract the phoneme duration. The duration for the word boundary symbol is set to zero. When input processing module UEI is used, the duration for tone/stress is zero.

We train the Fastspeech2 models with batchsize of 24 utterances on one GPU. We use the Adam optimizer [21] with  $\beta_1 = 0.9 $,  $\beta_2 = 0.98 $, and  $\epsilon = 10^{-9}$ and follow the same learning rate schedule in [22]. It takes 200k steps for training until convergence. We encourage readers to refer to \cite{ren2020fastspeech} for more training details.

The generated speech is represented as a sequence of 80-dim log-mel spectrogram frames, computed from 40ms windows shifted by 10ms. Waveforms are synthesized using a HifiGAN \cite{kong2020hifi} vocoder which generates 16-bit signals sampled at 16 kHz conditioned on spectrograms predicted by the TTS model. 

\subsection{Evaluation metrics}

We designed listening tests to evaluate the synthesized speech's naturalness (NAT) and speaker similarity (SIM) . Ten utterances were randomly chosen for evaluation in each scenario. Each utterance is rated by 14 listeners using the mean opinion score (MOS) on a five-point scale. Demos are available at  \url{https://haitongzhang.github.io/ipa-tts/}.





\begin{table*}[ht!]
    \caption{Naturalness (NAT) and similarity (SIM) MOS of synthesized speech by models with two different input processing modules.}
\centering
\begin{tabular}{ c c c c c c }
\hline\hline

\multirow{2}{*}{Speaker} &
\multirow{2}{*}{ \backslashbox{Model} {Text} } 

& \multicolumn{2}{ c }{CH}  & \multicolumn{2}{ c }{EN}\\
\cline{3-6}

&  & NAT & SIM  & NAT & SIM  \\
\hline

&  Ground-Truth  
&  $ 4.72 \pm 0.07 $  
&  $ 4.67 \pm 0.07 $  
&  $ 4.67 \pm 0.06 $  
&  $ 3.93 \pm 0.13 $  
  \\

&  Analysis-Synthesis
&  $ 4.62 \pm 0.09 $  
&  $ 4.62 \pm 0.08 $  
&  $ 4.60 \pm 0.08 $  
&  $ 3.90 \pm 0.14 $  
  \\
\hline

\multirow{2}{*}{CH} 

& MSEA 

&  $ 4.34  \pm 0.11 $   
&  $ 4.39  \pm 0.09 $  
&  $ 4.26  \pm  0.10 $ 
&  $ 1.99  \pm 0.11 $   
 \\

&  MUEI 
&  \textbf{ 4.35 $\pm$ 0.11 }  
&  \textbf{ 4.52 $\pm$ 0.08 } 
&   \textbf{ 4.37 $\pm$ 0.10  }
&  \textbf{ 2.09 $\pm$ 0.13 } 
 \\

\cline{1-6}

\multirow{2}{*}{EN} 

& MSEA 
&  $ 3.66 \pm 0.15 $    
& \textbf{  2.32 $\pm$ 0.14  }   
&  $ 4.39 \pm 0.09 $  
&   \textbf{ 3.83 $\pm$ 0.13  }
 \\

& MUEI 
&  \textbf{ 3.84 $\pm$ 0.13 }  
&   2.06 $\pm$ 0.11   
&  \textbf{ 4.44 $\pm$ 0.09 }  
&  3.75 $\pm$ 0.12 
 \\

\hline
\end{tabular}
\label{table:IPM}
\end{table*}

\section{Results and discussion}

\subsection{The impact of input processing modules} \label{IPM_res}

To study whether two different input modules impact the cross-lingual voice cloning performance, we trained two model variants using Dataset1: MSEA (the model with SEA) and MUEI (the model with UEI). 
The subjective evaluation results are provided in Table \ref{table:IPM}. It clearly shows that these two input processing modules have comparable performances on intra-lingual and cross-lingual voice cloning.




\subsection{Why fails cross-lingual voice cloning } \label{WHY_FAIL}

Table \ref{table:IPM} shows that the speaker similarity of CL VL is significantly lower than the intra-lingual performance. We learn from an informal listening test that many Chinese utterances synthesized using the English speaker's voice sound like the Chinese speaker and English utterances synthesized using the Chinese speaker's voice sound like the English speaker. In other words, only using IPA does not guarantee a perfect disentanglement between speaker identities and language symbols.

We hypothesize that this result can be attributed to the fact that (1) there are some non-overlapped IPA symbols across two target languages; (2) the suprasegmental, including tone and stress, are unique to only one of the target languages. To test our hypothesizes, we devised two input perturbation methods. 



\begin{itemize}
    \item \textbf{IPA perturbation:}  Replace all the IPA symbols in testing sentences in one language with the non-overlapped IPA symbols from the other language randomly. To remove the potential effect of tone/stress, we replace all tone/stress symbols with the special non-tone symbol.
    \item \textbf{Tone/stress perturbation:} Replace all tone symbols in Chinese testing sentences with the primary stress symbol in English, or replace all stress symbols in English testing sentences with the mid-tone in Chinese. To remove the potential effect of the non-overlapped IPA symbols, we replace them with their closest IPA symbols as in \cite{staib2020phonological}
\end{itemize}
 
 We use these two input perturbation methods to modify the original testing sentences and create in total six test datasets, namely CH and EN (original Chinese and English test data), CH\_IP and EN\_IP (Chinese and English test data with IPA perturbation), and CH\_TP and EN\_SP (Chinese and English test data with tone/stress perturbation). We then use MUEI to synthesize these six test datasets. We implement a speaker similarity preference test, where the raters are asked to judge whether the synthesized utterance is close to the voice of the Chinese speaker, the English one, or neither of them. Since using the proposed IPA or tone/stress perturbation may result in non-intelligible or accented speech, we ask the raters to focus on the speaker similarity during the test.The results are illustrated in Fig. \ref{fig:IPM_SPK_CH} and Fig. \ref{fig:IPM_SPK_EN}. 
 

\begin{table*}[t]
    \caption{Naturalness (NAT) and similarity (SIM) MOS of synthesized speech by models with different training data. }
\centering
\begin{tabular}{ c c c c c c c c  }
\hline\hline

\multirow{2}{*}{Speaker} &
\multirow{2}{*}{ \backslashbox{Model} {Text} } 

& \multicolumn{2}{ c }{CH}  & \multicolumn{2}{ c }{EN} & \multicolumn{2}{ c }{CS} \\
\cline{3-8}
&  & NAT & SIM  & NAT & SIM  & NAT & SIM  \\
\hline


&  Ground-Truth  
&  $ 4.72 \pm 0.07 $  
&  $ 4.67 \pm 0.07 $  
&  $ 4.67 \pm 0.06 $  
&  $ 3.93 \pm 0.13 $  
&  -
&  -
  \\

&  Analysis-Synthesis
&  $ 4.62 \pm 0.09 $  
&  $ 4.62 \pm 0.08 $  
&  $ 4.60 \pm 0.08 $  
&  $ 3.90 \pm 0.14 $  
&  -
&  -
  \\
   \hline

\multirow{4}{*}{CH} 

& C1E1 

&  $ 4.35  \pm 0.11 $   
&  $ 4.52  \pm 0.08 $  
&  $ 4.37  \pm  0.10  $ 
&  $ 2.09  \pm 0.13 $   
&  $ 3.72 \pm 0.14 $  
&  $3.74  \pm 0.11 $  
 \\

&  C1E4 
&  $4.44 \pm 0.10 $
&  $ 4.52 \pm 0.08 $
&  $ 4.02 \pm 0.14 $  
&  $2.25 \pm 0.12 $
&  $ 3.79 \pm 0.12 $  
&  $ 3.77  \pm 0.12 $  
 \\

 & C4E1 

&  \textbf{ 4.54  $\pm$ 0.10 }   
&  \textbf{ 4.62  $\pm$ 0.08 }  
&  $ 3.41  \pm  0.16 $
&  $ 2.60  \pm 0.13 $   
&  $ 3.89 \pm 0.12 $  
&  $ 3.96  \pm 0.12 $  
 \\

&  C4E4 
&  $4.49 \pm 0.09   $
&  $4.55 \pm 0.07 $
&   \textbf{ 4.07 $\pm$ 0.11 }
&  \textbf{ 4.17 $\pm$ 0.08 } 
& \textbf{ 4.06  $\pm$  0.12 }
&  \textbf{ 4.06  $ \pm$  0.11  } 
 \\

\cline{1-8}

\multirow{4}{*}{EN} 

& C1E1 
&  $ 3.84 \pm 0.13 $  
&  $ 2.06 \pm 0.11 $  
&  $ 4.44 \pm 0.09 $  
&  $ 3.75 \pm 0.12 $ 
&  $ 3.47 \pm 0.15 $  
&  $ 3.23 \pm 0.16 $  
 \\

& C1E4 
&  $ 3.26 \pm 0.16 $
&  $ 2.97 \pm 0.18 $ 
&  $ 4.41 \pm 0.09 $  
&  $ 3.81 \pm 0.11 $
&  $ 3.45 \pm 0.14 $  
&  $ 3.39  \pm 0.15 $  
 \\
 
 & C4E1 
&  $ 4.08 \pm 0.14 $  
&  $ 2.34 \pm 0.15 $  
&  $ 4.43 \pm 0.09 $  
&  $ 3.88 \pm 0.13 $ 
&  $ 4.08 \pm 0.12 $  
&  $ 3.59 \pm 0.14 $  
 \\

& C4E4 
&  \textbf{ 4.07 $\pm$ 0.11 }  
&  \textbf{ 3.68 $\pm$ 0.14 }  
&  \textbf{ 4.46 $\pm$ 0.09 }  
&  \textbf{ 3.98 $\pm$ 0.13 } 
&  \textbf{ 4.11 $\pm$ 0.13 } 
&  \textbf{ 3.63 $\pm$ 0.14  }
 \\

\hline
\end{tabular}
\label{table:NUM_MOS}
\end{table*}

 \subsubsection{The effect of non-overlapped IPA}
\begin{figure}[t]
    \centering
    \includegraphics[width=0.4\textwidth,height=0.25\textwidth]{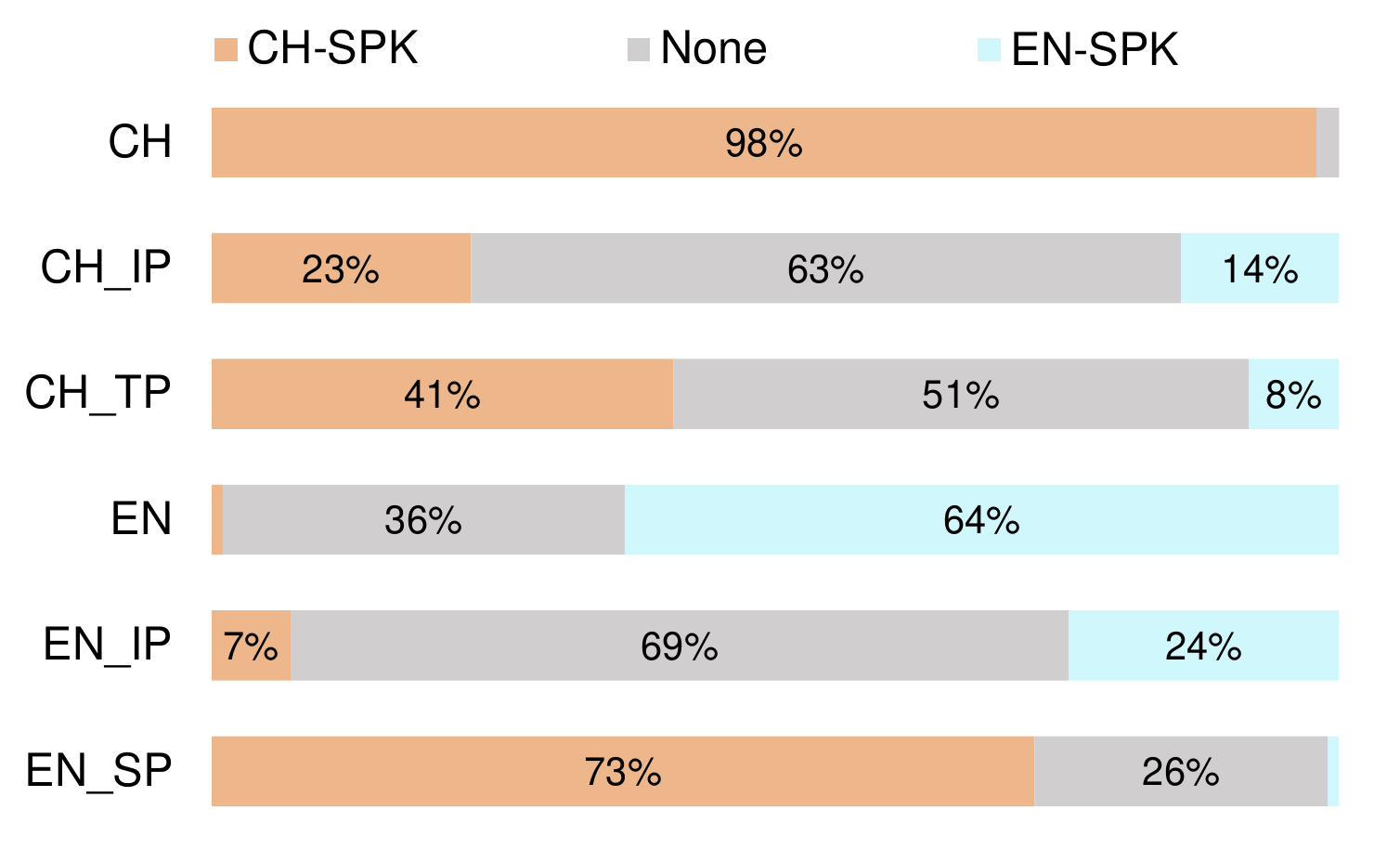}
    \caption{\emph{Speaker similarity preference of synthesized utterances of six test datasets using the Chinese speaker's voice.}}
    \label{fig:IPM_SPK_CH}
\end{figure}

\begin{figure}[t]
    \centering
    \includegraphics[width=0.4\textwidth,height=0.25\textwidth]{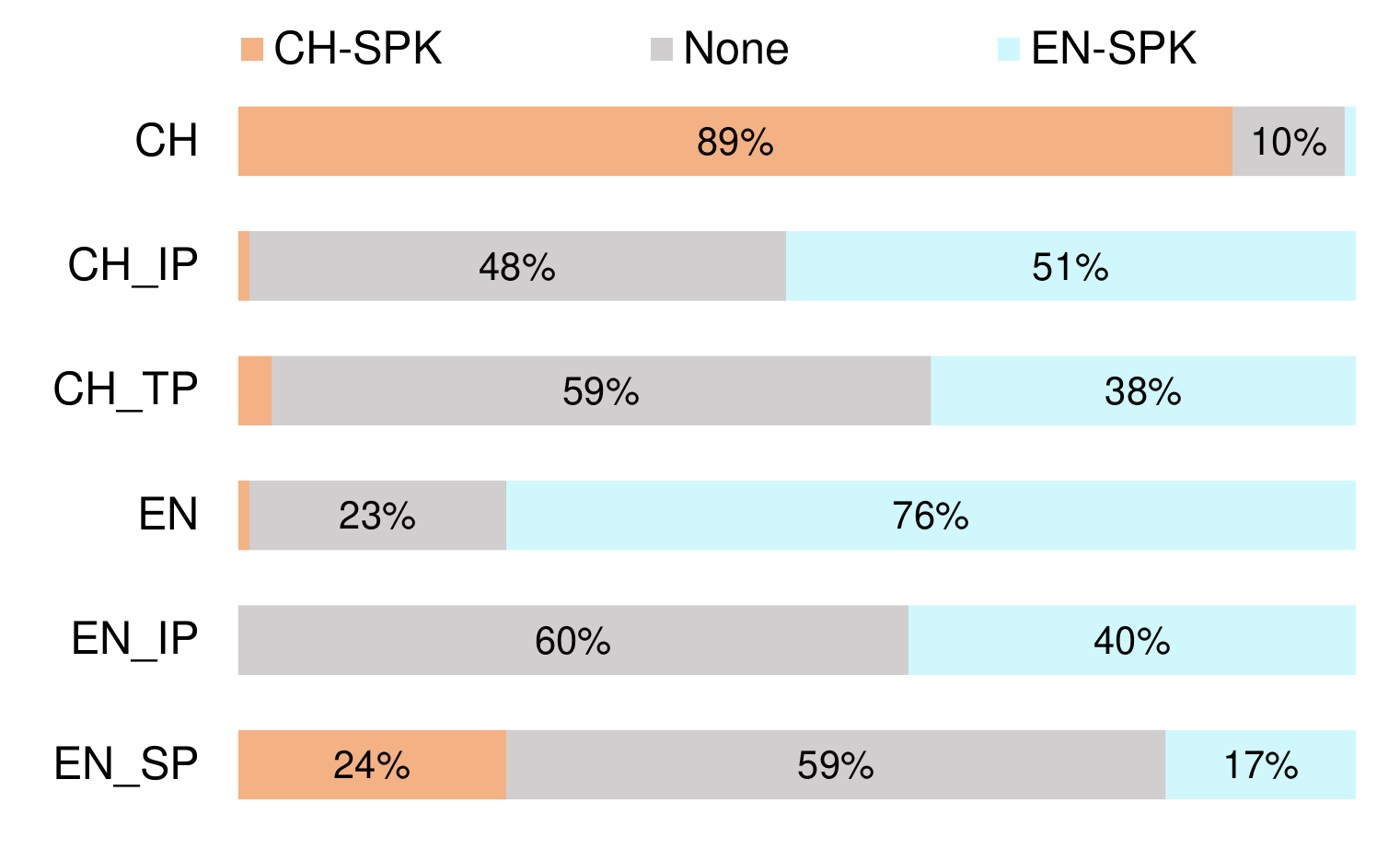}
    \caption{\emph{Speaker similarity preference of synthesized utterances of six test datasets using the English speaker's voice.}}
    \label{fig:IPM_SPK_EN}
\end{figure}

As shown in Fig. \ref{fig:IPM_SPK_CH} and Fig. \ref{fig:IPM_SPK_EN}, with IPA perturbation, the speaker similarity of the Chinese synthesized utterances decreases significantly for the Chinese speaker, and increases significantly for the English speaker (see CH\_IP). When using IPA perturbation to the English text, the speaker similarity for the Chinese speaker increases while that for the English speaker decreases (see EN\_IP). These results support our hypothesis that the non-overlapped IPA symbols are likely to contain some speaker information. 



 \subsubsection{The effect of tone/stress} 
 With tone perturbation, the speaker similarity of the Chinese synthesized utterances decreases significantly for the Chinese speaker and increases significantly for the English speaker (see CH\_TP). This indicates that the stress symbols in English contain speaker information of the English speaker. For the English text, stress perturbation significantly increases the speaker similarity for the Chinese speaker, while it decreases the speaker similarity for the English speaker by a large margin (see EN\_SP). This reveals that the tone symbols in Chinese are also responsible for the speaker information leakage.




       

       

       

       

       
       
       


\subsection{The number of speakers}
In section \ref{WHY_FAIL}, we find that both non-overlapped (or language-unique) IPA and tone/stress symbols are likely to contain some speaker information, which causes the model to fail cross-lingual voice cloning. In this section, we continued the investigation by proposing the following hypothesizes. 

\textbf{Hypothesis 1:} \textit{  The secondary or indirect reason our models fail CL VL is that we only use two speakers as training data. In other words, as we increase the number of speakers, this failure can be avoided.}

\textbf{Hypothesis 2:}  \textit{  Increasing the number of speakers in only one language would result in success to CL VL for speakers in this language, but a failure for the speaker in the other language.}



To test our hypothesizes we compared several model variants trained with different subsets of Dataset2:


\begin{itemize}
    \item C1E1: Model trained with one Chinese speaker and one English speaker (MUEI in section \ref{IPM_res} ).
    \item C1E4: Model trained with one Chinese speaker and four English speakers.
    \item C4E1: Model trained with four Chinese speakers and one English speaker.
    \item C4E4: Model trained with four Chinese speakers and four English speakers.
\end{itemize}

 We use the input processing module UEI in this scenario for fair comparisons. The results of naturalness and speaker similarity MOS evaluations are illustrated in Table \ref{table:NUM_MOS}.

As shown in Table \ref{table:NUM_MOS}, the speaker similarity of cross-lingual voice cloning tends to increase as the number of speakers increases. In addition, when only increasing the number of speakers in one target language (i.e., C1E4 or C4E1), the speaker similarity improvement on CL VL for speakers in that target language is more significant than speakers in the other language. However, the naturalness MOS for speakers in that target language shows a decreasing trend. We suspect that the models have learned to disentangle the speaker identities from the language-unique symbols but fail to synthesize natural cross-lingual speech due to the imbalanced distribution of the training data. Hence, increasing the number of speakers in all languages provides the best CL VL performance.

Furthermore, we provided the results of code-switched synthesized speech. Model C1E1 performs a decent performance on code-switched utterances. We suspect that when synthesizing these code-switched sentences, the non-overlapped IPA symbols in two languages are likely to compete with each other to leak the speaker information, and tones in Chinese and stresses in English would join the competition as well. As the results indicate, they usually play a tie game in that they fail to leak the speaker information, and the speaker embedding plays its full role. In addition, we observed a steady improvement through increasing the number of speakers, and Model C4E4 achieves the best code-switched performance.

\begin{figure}[t]
    \centering
    \includegraphics[width=0.5\textwidth,height=0.4\textwidth]{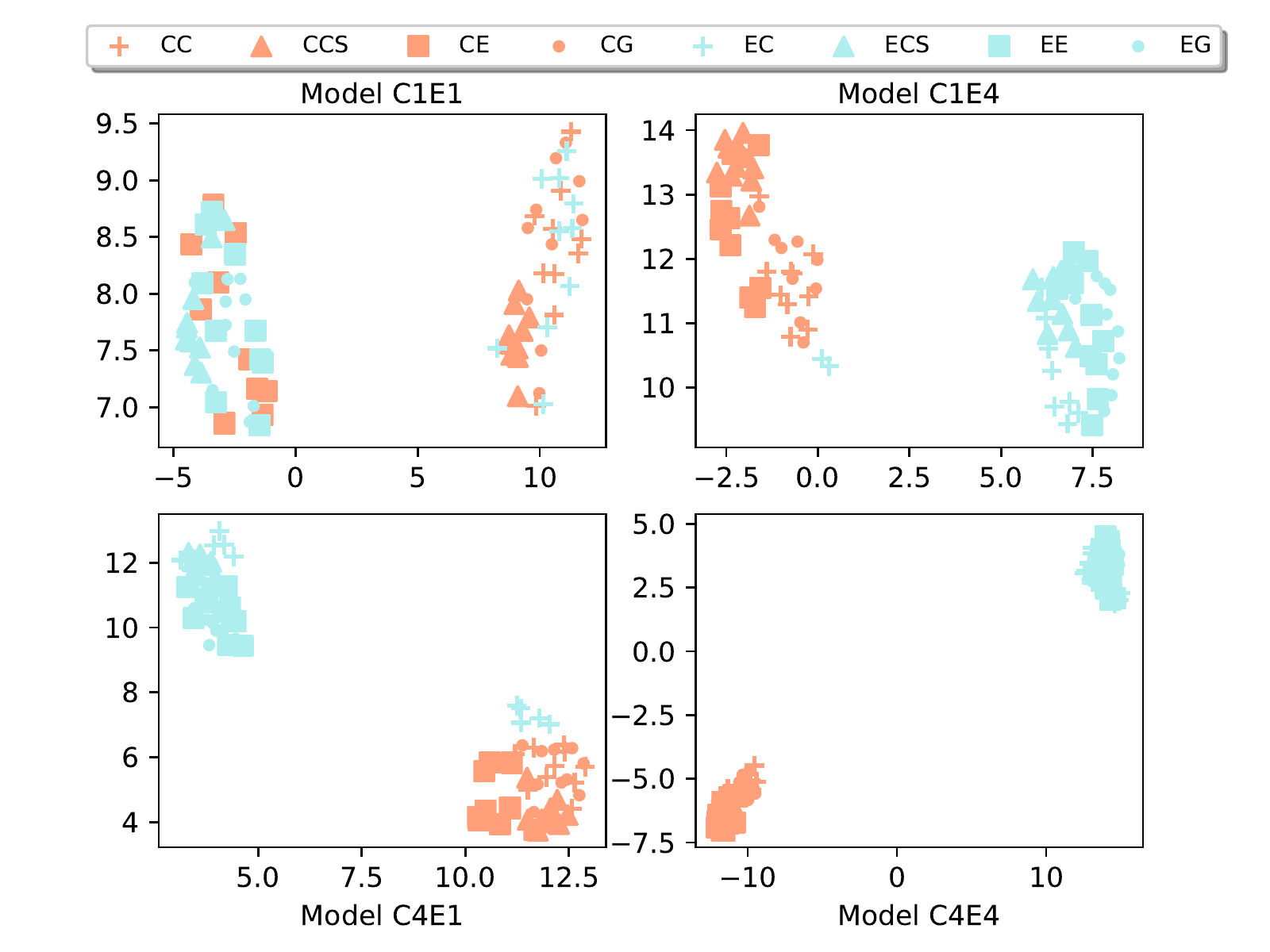}
    \caption{ \emph{Visualizing the effect of the number of speakers, using 2D UMAP \cite{mcinnes2018umap} of speaker embeddings\cite{wan2018generalized} computed from speech synthesized with different speaker and text combinations. The orange color represents speech from the Chinese speaker, while blue represents the English one. $+$ , $\triangle$, and $\Box$  denotes the Chinese, code-switched, and  English text, respectively; and $\circ$ refers to the ground truth utterances.}}
    \label{fig:IPM_SPK_CH_plot}
\end{figure}

Fig. \ref{fig:IPM_SPK_CH_plot}  provides a vivid visualization of the speaker similarity of synthesized speech as well. Model C4E4 provides the ideal clustering: the synthesized speech clusters close to the respective target speaker's ground-truth utterances, and the synthesized speech from two speakers is separated at a considerable distance.

\section{Conclusions}

In this study, we present an empirical study of building an IPA-based cross-lingual non-autoregressive TTS model. We conclude our findings as follows.
\begin{itemize}

    \item  The way to process IPA and tone/stress sequences has a negligible impact on the CL VL performance.
    
     \item IPA alone does not guarantee successful CL VL performance since the language-unique IPA and tone/stress symbols are likely to leak the speaker information.
     
    \item One simple but effective method to improve the CL VL performance of IPA-based CL TTS is to increase the number of speakers in all languages.
    
\end{itemize}

Although our findings are based on the non-autoregressive TTS model, they should be generalized well to other TTS frameworks.


\bibliographystyle{IEEEtran}

\bibliography{refs}

\end{document}